\documentclass[journal]{IEEEtran}
\usepackage{cite}
\usepackage{subfig}
\usepackage{float}
\usepackage{graphicx}
\usepackage{amsmath}
\DeclareMathOperator*{\argmax}{\arg\,\max}

\usepackage[colorlinks]{hyperref}
\hypersetup{linkcolor=blue, citecolor=red, urlcolor=teal}
\begin{document}
\title{A Digital Fuzzy Edge Detector for Color Images}
\author{Yuan-Hang Zhang, Xie Li, Jing-Yun Xiao
\\\emph{Department of Computer Science and Technology}\\\emph{University of Chinese Academy of Sciences}\\Beijing 100049, China
\thanks{The authors are with the University of Chinese Academy of Sciences, Beijing 100049, China (e-mail: zhangyuanhang15, lixie15, xiaojingyun15@mails.ucas.ac.cn).}
\thanks{Manuscript received Jan 12, 2017.}}

\maketitle

\begin{abstract}
Edge detection is a classic problem in the field of image processing, which lays foundations for other tasks such as image segmentation. Conventionally, this operation is performed using gradient operators such as the Roberts or Sobel operator, which can discover local changes in intensity levels. These operators, however, perform poorly on low contrast images. In this paper, we propose an edge detector architecture for color images based on fuzzy theory and the Sobel operator. First, the R, G and B channels are extracted from an image and enhanced using fuzzy methods, in order to suppress noise and improve the contrast between the background and the objects. The Sobel operator is then applied to each of the channels, which are finally combined into an edge map of the origin image. Experimental results obtained through an FPGA-based implementation have proved the proposed method effective.
\end{abstract}


\section{Introduction}
\IEEEPARstart{E}{dge} detection plays a fundamental role in the field of image processing. An \emph{edge} is a collection of connected pixels where the intensity level changes abruptly.\cite{Gonzalez} Edges are especially important for image recognition and analysis since they can capture local features and provide useful information.

So far, techniques for edge detection have been studied extensively. Edges can usually be found in parts of an image where transition occurs, either between different objects, different regions, or between objects and the background. In this view, gradients are effective descriptors of edges. Consequently, various gradient-based edge detectors were proposed during the 1960s, including the Robert operators, Robinson operators, Prewitt operators, and Sobel operators, first suggested by Irwin Sobel and Gary Feldman\cite{Sobel68}. A more accurate detector known as the \emph{Canny detector} was later proposed by J.~Canny, which smooths the input image with a Gaussian filter and applies non-maxima suppression for optimal results.\cite{Canny86}

For digital images, derivatives can be approximated with discrete differentiation. Therefore, first-order edge detectors are easy to implement and widely used. Among them, the Sobel operators are especially preferred because they are non-linear filters with image smoothing, and thus can produce less fragmentary edge images.

Nevertheless, early implementations of Sobel operators, mainly software-based and DSP-based were slow and unable to support real-time operations. The recent introduction of Field Programmable Gate Arrays (FPGAs) have succesfully resolved the problem. An FPGA is a device that enables real-time parallel processing with pipeline technology, which in previous studies were used to implement Sobel detectors\cite{Guo10}. Experimental results in \cite{Guo10} proved the design an effective approach to real-time edge detection. A novel pipelined structure was also proposed to speed up the operations\cite{Abbasi07}.

However, when Sobel operators are applied to low-contrast images, the results are often less satisfactory, with more isolated points and significant distortions. To this end, we introduce \emph{fuzzy theory} to enhance the edges before applying the Sobel operators. This method was studied in \cite{Suryakant12} and \cite{Zhang15}. In this paper, we combine the fuzzy method and the Sobel edge detector to implement a FPGA-based color image detector with improved accuracy for low-contrast images.

The rest of the paper is arranged as follows. The Sobel edge detection filter is introduced in Section~\ref{sec2}. Theoretical foundations for the preprocessing procedures are elaborated in Section~\ref{sec3}. Section~\ref{sec4} gives the FPGA-based hardware implementation for this paper. In Section~\ref{sec5}, we present the experimental results. Future prospects are briefly discussed in Section~\ref{sec6}.

\section{The Sobel Edge Detection Filter}\label{sec2}
The Sobel edge detection filter is a commonly used edge detector that computes an approximate gradient of the image intensity function. For each pixel in the image, it obtains the vertical and horizontal components of the gradient by applying convolution with two 3-by-3 masks defined as:
\[H_1=\begin{bmatrix}
-1&-2&1\\0&0&0\\1&2&1
\end{bmatrix},\]
\[H_2=\begin{bmatrix}
-1&0&1\\-2&0&2\\-1&0&1
\end{bmatrix}.\]
Specifically, the horizontal and vertical gradients of the pixel $(p,q)$ can be written as
\begin{align*}
G_{x}(p,q) = &~G(p+1,q-1)+2G(p+1,q)
\\&+G(p+1,q+1)-(G(p-1,q-1)
\\&+2G(p-1,q)+G(p-1,q+1)),
\\G_{y}(p,q) = &~G(p-1,q+1)+2G(p,q+1)
\\&+G(p+1,q+1)-(G(p-1,q-1)
\\&+2G(p,q-1)+G(p+1,q-1)),
\end{align*}
It can be shown that the coefficient $2$ endows the operator with noise suppression capabilities.\cite{Gonzalez} The $L^2$ norm of the corresponding gradient vector can then be computed by
\begin{equation}
G = \sqrt{G_{x}^2 + G_{y}^2}\label{sobel-l2norm}
\end{equation}
However, Eq.~\eqref{sobel-l2norm} is of high computational costs, and the evaluation of square roots will require extra modules such as a CORDIC unit\cite{Bailey}, rendering the circuit complex. Thus the $L^1$ norm is frequently used to approximate this magnitude:
\begin{equation}
G \approx |G_x| + |G_y|
\end{equation}
An improved version of the original Sobel operator uses two extra convolution kernels that detect gradients along the diagonals:
\[H_3=\begin{bmatrix}
-2&-1&0\\-1&0&1\\0&1&2
\end{bmatrix},\]
\[H_4=\begin{bmatrix}
0&1&2\\-1&0&1\\-2&1&0
\end{bmatrix}.\]
For the sake of simplicity, we shall confine ourselves to the original Sobel operator with only two masks.

\section{Preprocessing}\label{sec3}
\subsection{The Otsu Method}
For this study, threshold selection is realized using \textit{the Otsu Method}. This adaptive method is considered optimal in the automatic evaluation of a suitable threshold. The algorithm assumes two categories of pixels, object pixels and background pixels. The algorithm finds a threshold $g$ such that the mean grayscale of object, $\mu_0(g)$, and the mean grayscale of the background, $\mu_1(g)$ are furthest from the mean grayscale of the entire image, $\mu_g$. This threshold $g$ can be found by maximizing between-class variance. 

Its procedure can be described as follows.

\[p(k)=\frac{1}{MN}\sum_{g(i,j)=k}1\]
where $p(k)$ is the frequency of the pixels in the picture and $g(i,j)$ is the gray-level of point $(i,j)$ in the original image.
The percentage of the object is
\[\omega_0(t)=\sum_{0\leq i\leq t}p(i),\]
and the percentage of the background is
\[\omega_1(t)=\sum_{t< i\leq m-1}p(i).\]
The means of object $\mu_0(t)$ is 
\[\mu_0 (t) = \sum_{0\le i\le t}\dfrac{ip(i)}{\omega_0(t)}\]
and the mean of background $\mu_0$ is 
\[\mu_1(t) = \sum_{t<i\le m-1} \dfrac{ip(i)}{\omega_0(t)}\]
Therefore, the overall mean is 
\[\mu = \omega_0(t)\mu_0(t) + \omega_1(t)\mu_1(t) \] 
The final threshold value $g$ is 
\begin{equation}
g = \argmax_{0\le t \le m-1} (\omega_0 (t)(\mu_0(t) - \mu)^2 + \omega_1(t) (\mu_1(t) - \mu)^2)\label{OtsuTh}
\end{equation}
Intuitively, this formula gives the maximum between-class variance, which can roughly be viewed as the contrast between two classes of pixels, black and white. The larger the variance is, the better the contrast will be. Therefore, the optimal segmentation threshold $g$ for the original image found by Eq.~\eqref{OtsuTh} splits the image into object and background effectively.

This method can be realized in an FPGA fabric using the fixed-point number system and supplementary modules.\cite{Pandey14}
\subsection{Fuzzy Sets for Edge Detection Enhancement}
We introduce fuzzy theory to enhance the edge detection capabilities of the Sobel detector. The theory of fuzzy sets was proposed by Zadeh in 1965 and has gained much weight in image processing. It uses the degree of membership function to determine whether a given point is ``black" or ``white", a concept that is usually subjective and hard to accurately describe.

Formally, a \textit{fuzzy set} $B$ in a finite set $X = \{x_1,x_2,\ldots, x_n\}$ is represented as $B = \{(x,\mu_A(x))\mid x\in B\}$, where $\mu_B(x): B \rightarrow [0,1]$ is the function that represents the belongingness or degree of membership in of an element in $x$ in the finite set. In this paper, we only have two fuzzy sets $B$ and $W$ ($B$ for black, $W$ for white), thus the non-belongingness of an element of set $B$ is $1-\mu_B(x)$, which is also the belongingness of fuzzy set $W$. Note that the degree of memebership function is usually a family of functions determined by the threshold parameter, which in this case is obtained via the Otsu method.

Some commonly used Common degree of membership include $r$ distribution, Cauchy distribution, Gauss distribution, and the S-shaped distribution. 

The Cauchy distribution is defined as
\[\mu_A(x) = \dfrac{1}{1+\alpha(x - c)^\beta}\]
where $\alpha > 0 $, $\beta$ is a positive even integer, and $c$ is the center of distribution.

Another version of this distribution is
\[\mu_A(x) = \dfrac{1}{1+\alpha_1|x-c|^{\beta_1}}\]
where $\alpha_1,\beta_1>0$.

The S-shaped distribution is
\[f(x)=\left\{
\begin{aligned}
&0,\qquad  &x\le a\\
&a(\dfrac{x-a}{c-a})^2,\qquad &a<x\le b\\
&1-a(\dfrac{x-c}{c-b})^2,\qquad &b<x\le c\\
&1, \qquad &x>c
\end{aligned}
\right.
\]

To employ fuzzy theory in image segmentation and edge detection, one approach is to divide an image into multiple layers by different thresholds, which can can be obtained by the maximum fuzzy entropy\cite{Zhang15}. Another way is to directly transform the original image into an enhanced version using the fuzzy function. In this study, we select a universal polynomial function
\begin{align*}-2.0454098505641*10^{-7}x^4+7.615967514125*10^{-5}x^3
\\-0.0041249658333x^2+0.4911541875107x,
\end{align*}
where $x$ is the original gray value. This function is independent to the threshold and similar to the ``increase contrast" function used in Adobe Photoshop.

The preprocessing procedure carried out in the experiment can be summarized as follows:
\begin{enumerate}
\item Calculate the threshold $T$ using the Otsu method.
\item Apply threshold $T$ to get the degree of membership $\mu_W$.
\item Obtain the fuzzy image from the function
\begin{align*}
f: [0,1]&\rightarrow [0,255]
\\x &\mapsto 255\cdot \mu_W(x).
\end{align*}
\end{enumerate}
\section{Hardware Implementation}\label{sec4}
\begin{figure*}
\includegraphics[width=180mm]{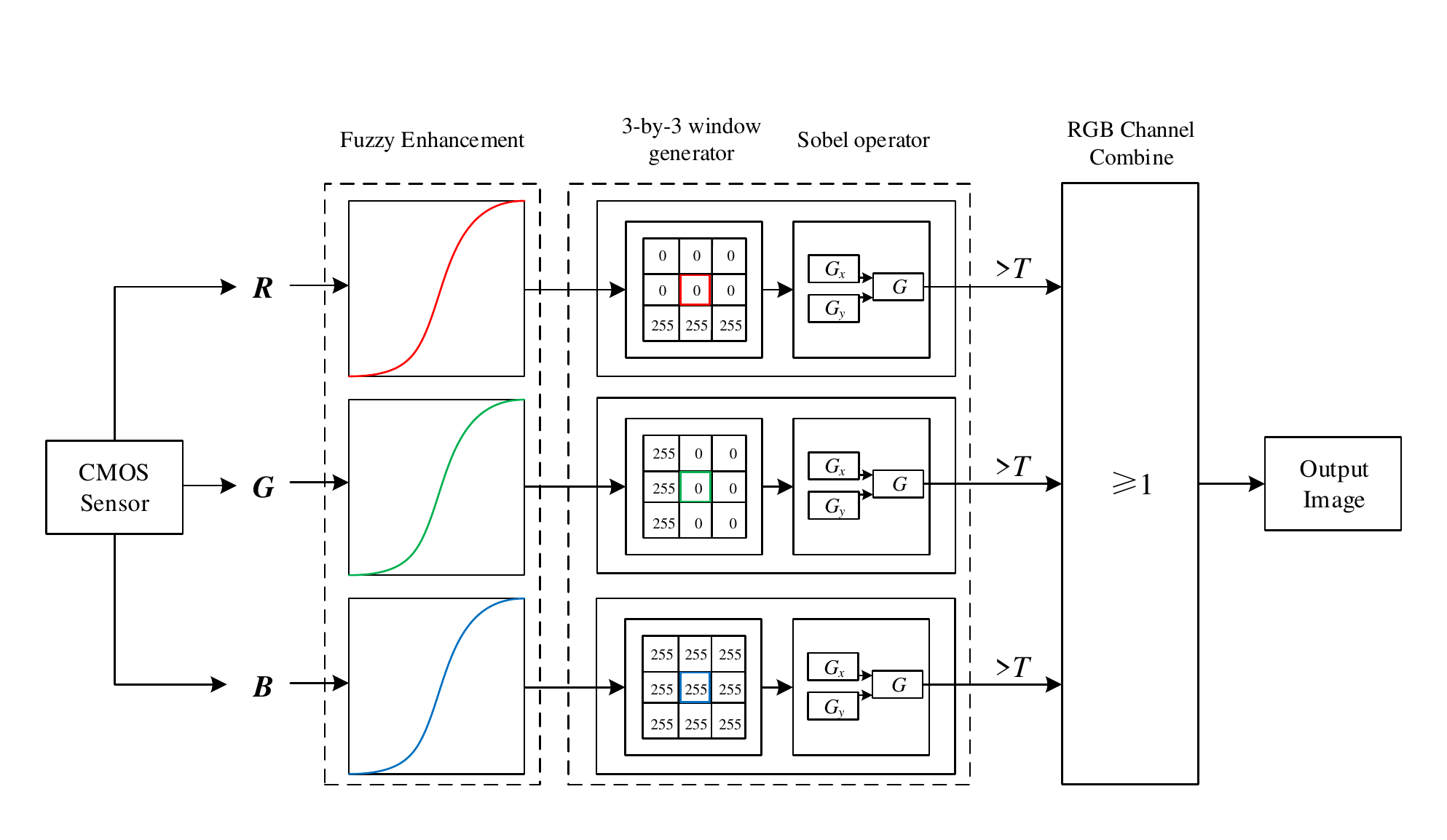}
\caption{General architecture of our fuzzy edge detection system.}\label{fig1}
\end{figure*}

The proposed architecture is illustrated in Fig.~\ref{fig1}. This system is composed of three modules, the fuzzy preprocessor, the window generator and the Sobel operator.
\subsection{Fuzzy preprocessor}
Fuzzy preprocessor is the module that enhances the original image to increase contrast between dark and bright regions. The data input of this module are the individual pixels from the original image. For this experiment, the polynomial degree of membership function is hard-coded into this module. The resulting image is then stored and accessed by the window generator. The final threshold is determined by the Otsu method using a separate program, although a complete FPGA solution is also possible\cite{Pandey14}.
\begin{figure}[H]
\centering
\includegraphics[width=88mm]{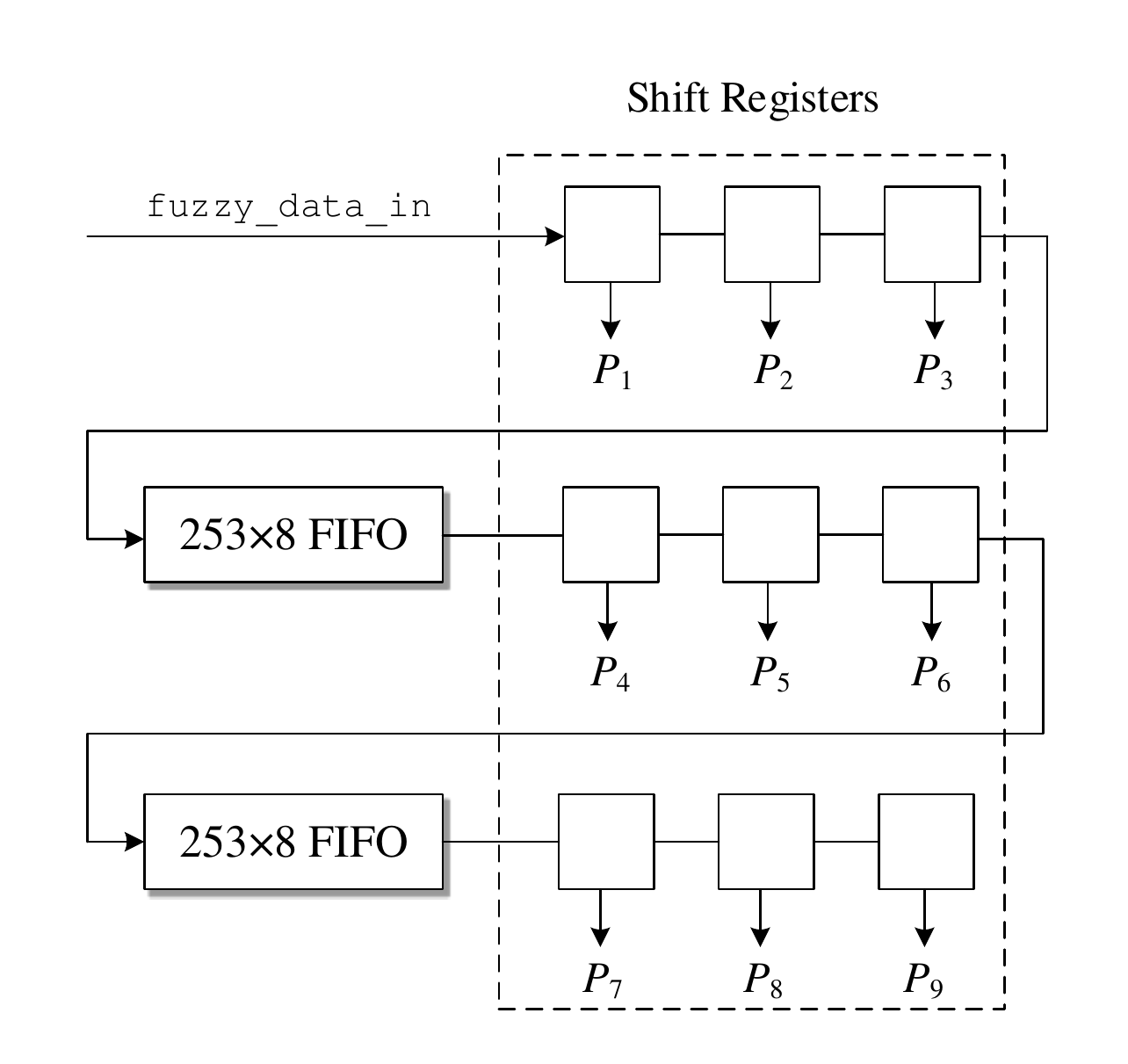}
\caption{Generating 3-by-3 windows from the input stream.}\label{fig2}
\end{figure}
\subsection{Window generator}
Since the image input is in the form of a data stream, only one pixel can be transmitted during each clock cycle. To obtain the nine necessary pixels simultaneously, we deploy three rows of shift registers and two FIFOs to buffer the data. The clock signal \texttt{clk} controls the input so when the nine 8-bit arrays $P_1,P_2,\cdots, P_9$ are full, a 3-by-3 window is prepared for the Sobel algorithm. The two $253\times 3$ FIFOs are generated by RAMs.

The structure of the window generator is shown in Fig.~\ref{fig2}. It consists of two FIFOs with length of 253 and width of 8 and three 3-bit shift register. The three shift registers ``crops" a 3-by-3 window, while the two FIFOs are used to buffer the data between while traversing different rows of the image matrix, which has length $256-3 = 253$. When we shift the data line, the position of the window is also shifting.
\subsection{Sobel operator}
The function of the Sobel operator has been explained in Sec.~\ref{sec2}. The inputs of this module are the nine entries of the window matrix marked $P_1$ to $P_9$, and the output is either 1 or 0. 1 indicates that the pixel we are calculating is an edge pixel and 0 otherwise. The result is obtained by thresholding the gradient, given by the sum of absolute magnitudes of the two components.

It is worth noticing that the operations can be effectively reduced by half if the additions are grouped properly. We define the following quantities, first proposed in \cite{Liu81}:

\textit{Neighbouring sums}:
\[ \mathbf{NS}_r = G(i,j)+G(i+1,j),\]
\[ \mathbf{NS}_c = G(i,j)+G(i,j+1).\]

\textit{Partial sums}:
\[\mathbf{PS}_r = G(i,j)+2G(i+1,j)+G(i+2,j),\]
\[\mathbf{PS}_c = G(i,j)+2G(i,j+1)+G(i,j+2.)\]

\textit{Interlaced differences}:
\begin{align*}
\mathbf{ID}_r =& G(i,j)+2G(i+1,j)+G(i+2,j)
\\&-(G(i,j+2)+2G(i+1,j+2)+G(i+2,j+2)),
\end{align*}
\begin{align*}
\mathbf{ID}_c =& G(i,j)+2G(i,j+1)+G(i,j+2)
\\&-(G(i+2,j)+2G(i+2,j+1)+G(i+2,j+2)).
\end{align*}
The original procedure described in Section~\ref{sec2} requires 15 summations to compute the gradient for a single pixel. As the 3-by-3 window shifts with the data stream, it is clear that some pixels must be accessed more than once. Now, if we save the above intermediate results, the process can be accelerated. First, add the gray value of each pixel with its right neighbour and obtain a matrix for right-neighbour sum (the same procedure is repeated on the columns). Then, add each entry of the obtained matrix with its right entry as same as the first step, matrix for partial sum is obtained. Last, compute the interlaced difference for every pixel, the absolute value of interlaced difference is exactly the horizontal gradient. In this manner only seven sums are needed to compute the gradient for each pixel, accelerating the operations significantly.
\section{Experimental Results and Analysis}\label{sec5}
\begin{figure*}[!t]
\centering
\subfloat[Pens, ground truth]{\includegraphics[width=1.4in]{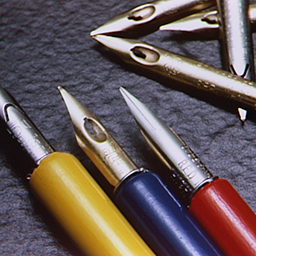}%
\label{The Pens image}}
\subfloat[R channel]{\includegraphics[width=1.4in]{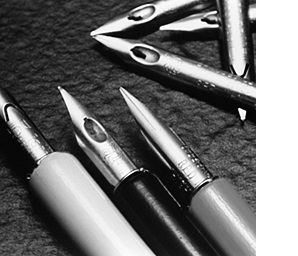}%
\label{R}}
\subfloat[G channel]{\includegraphics[width=1.4in]{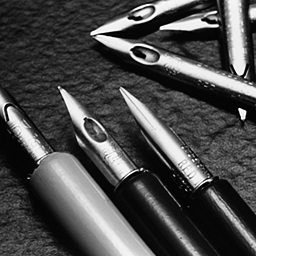}%
\label{G}}
\subfloat[B channel]{\includegraphics[width=1.4in]{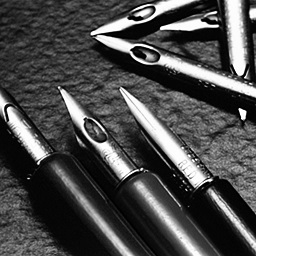}%
\label{B}}
\caption{The original test picture.}
\label{fig_sim}
\end{figure*}

\begin{figure*}[!t]
\centering
\subfloat[Threshold = 400, fuzzy preprocessing enabled. (i) Edge map. (ii), (iii), (iv) R, G, and B channel edge maps.]{\includegraphics[width = 6in]{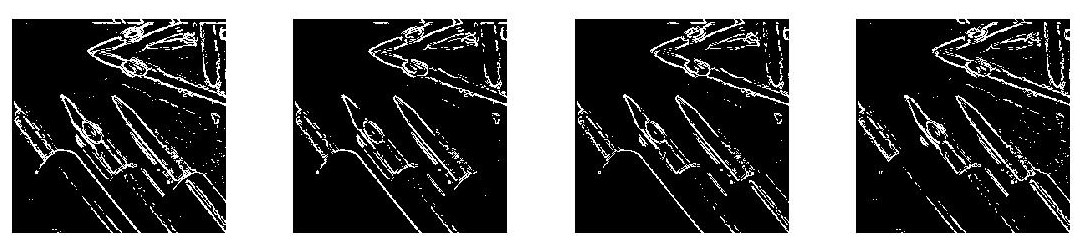}%
\label{fig_pen_fuzzy}}\hfill
\subfloat[Threshold = 400, fuzzy preprocessing disabled. (i) Edge map. (ii), (iii), (iv) R, G, and B channel edge maps.]{\includegraphics[width = 6in]{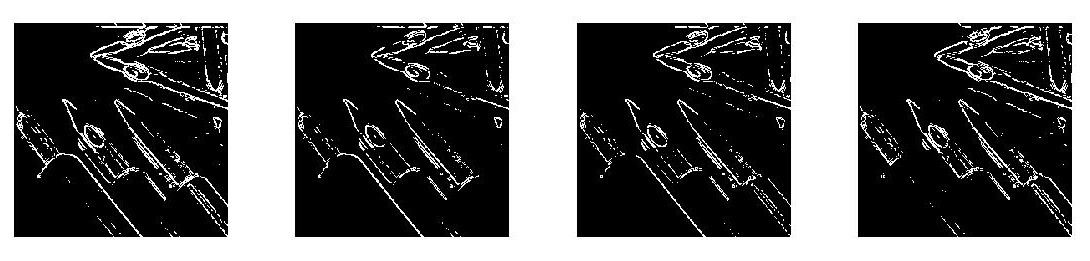}%
\label{fig_pen_non_fuzzy}}\hfill
\caption{A comparison of fuzzy method and non-fuzzy edge detectors.}
\label{fig4}
\end{figure*}

The circuit design is completed for $256\times 256$ 8-bit RGB images in Verilog HDL and simulated in ModelSim 10.1c. Since CMOS sensors such as OV9650 can be configured to output in RGB mode, for test purposes we store the three channels of an image in three separate text files to be read by the testbench program. The eight images in Fig.~\ref{fig4} are the corresponding edge maps of Fig.~\ref{fig_sim}, obtained using our edge detector.

We compare the results of standalone Sobel operators to that obtained with fuzzy preprocessing enabled. Under the same threshold of the Sobel operator, it is obvious that our method is able to extract more details from the original picture. In Fig.~\ref{fig4}, the edge detector using non-fuzzy method produces incomplete edges of the pens. On the other hand, with fuzzy preprocessing the different regions become more distinguished, so more edges are detected, as desired.

The reason fuzzy method produces this more ideal outcome is that fuzzy method increases the contrast of different portions of the image and thus increases the difference of the pixels on either side of the edge. However the noise level is not significantly reduced as expected. For example, at the upper-left corner of Fig.~\ref{fig4}(a), while brightness level fluctuations in this area are clearly visible, edges are barely detected.

\section{Future Work}\label{sec6}
\subsection{Noise Reduction}
Several solutions to resolve the noise problem mentioned in Section~\ref{sec5} are possible. For instance, as is shown in \cite{Zhang15}, the Canny operator can produce an ideal outcome. Second, smoothing operation plays an important role in de-noising. It is feasible to add a smoothing convolution module before the Sobel module to reduce the noise level. Third, in this experiment we used a simple polynomial function for fuzzy preprocessing for simplicity. A more appropriate fuzzy transformation function has yet to be found.

\subsection{An Introduction to Memristor Technology}
Although the edge detection based on Sobel operator and fuzzy set has been thoroughly studied, we wish to extend and optimize our work using memristors.

In 1971, Chua proposed the possibility of a new electronic component called a memristor. In 2008, Hewlett-Packard implemented the first instance of memristor which could be used as a basic component of a digital circuit. Memristors are capable of performing logic operations with its unique memristive features. Moreover, the memory of memristor is non-volatile to the loss of voltage. Furthermore, the voltage and resistance features can be used to compute complex logic functions\cite{Zhou13}. The idea of conventional edge detection based on memristor is building XOR function to test the edge in the image using memristor \cite{Merrikh14}. Moreover, a memristor-based nonvolatile latch design was proposed in\cite{Robinett10}, which greatly reduces the size of latch circuit. One future prospect for edge detection is to implement a Sobel edge detector using memristors. We believe that the size of the circuit will be reduced significantly and the whole circuit will be more non-volatile to voltage fluctuations. 
\begin{figure}[H]
\centering
\includegraphics[width=88mm]{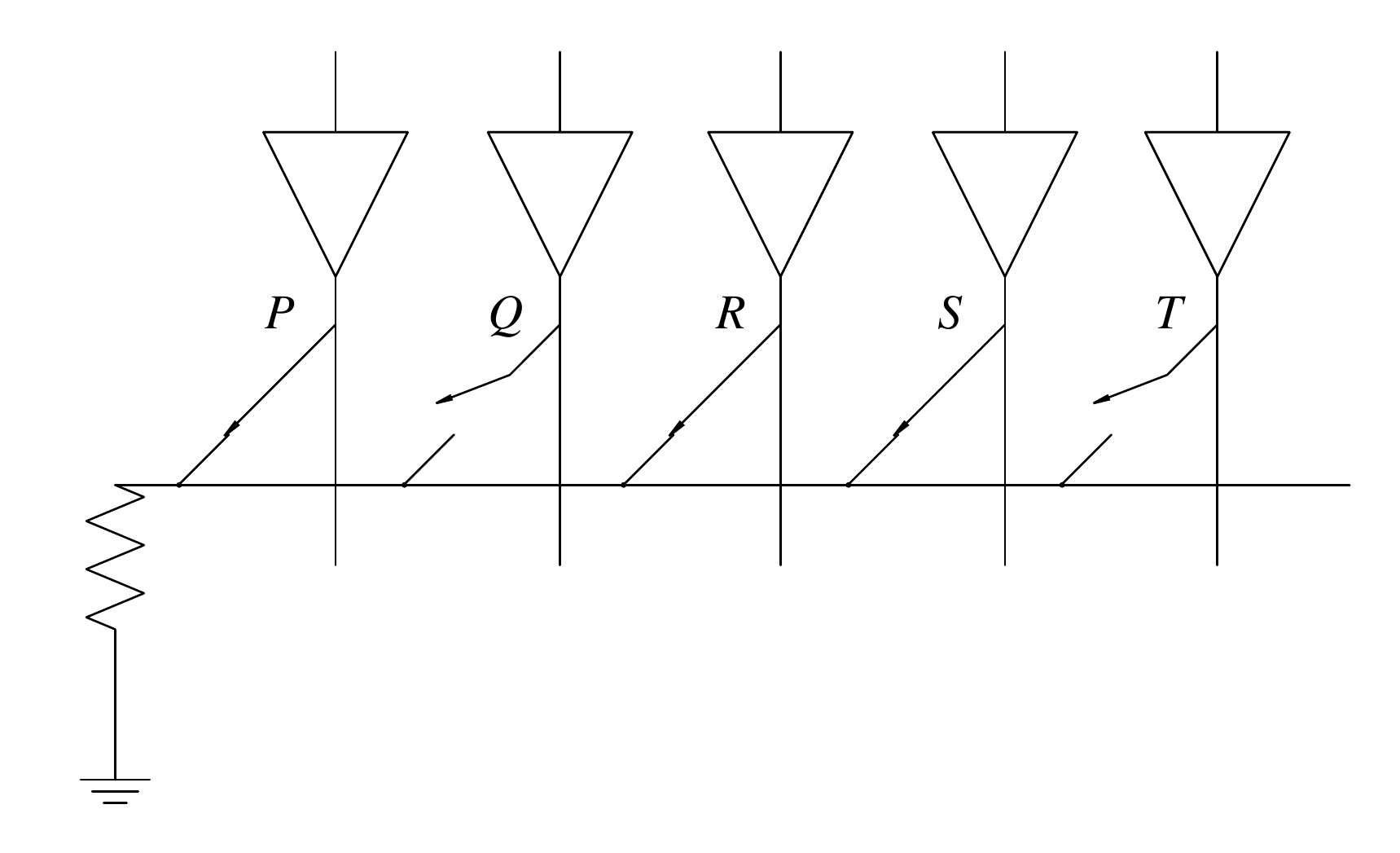}
\caption{The structure of an XOR-based memristor.\cite{Zhou13}}\label{fig5}
\end{figure}
\begin{center}

\begin{tabular}{|c|c|c|c|c|c|c|}
\hline
Step & Operation & $V_p$ & $V_q$ & $V_r$ & $V_s$ & $V_t$\\
\hline
\hline
1 & $r = q \rightarrow r$ &  & $V_{\rm{cond}}$ & $V_{\rm{set}}$& &  \\
\hline 
2 & $s = p\rightarrow s$ & $V_{\rm{cond}}$ & & &$V_{\rm{set}}$ & \\
\hline
3 & $t = r \rightarrow t$ & & & $V_{\rm{cond}}$ & & $V_{\rm{set}}$\\
\hline
4 & $t = s\rightarrow t$ & & & & $V_{\rm{cond}}$ &$V_{\rm{set}}$\\
\hline

\end{tabular}

Table 1: Stimulation sequence for the XOR function.
\end{center}

\begin{figure}[H]
\centering
\includegraphics[width=88mm]{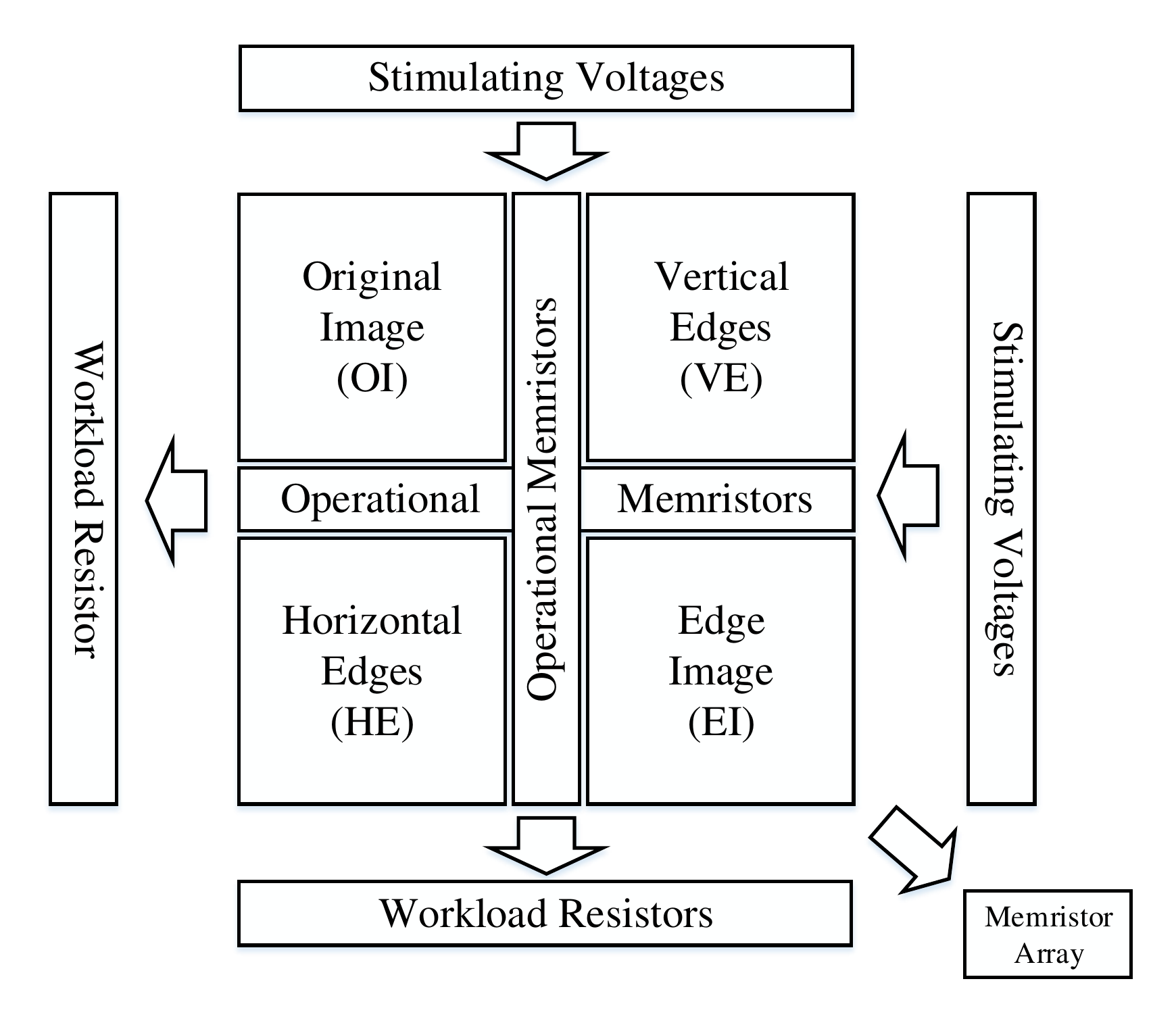}
\caption{The architecture of a memristive edge detector.\cite{Zhou13}}\label{fig6}
\end{figure}

In \cite{Zhou13}, an edge detector based on memristor was proposed. A memristor has two states, high impedance and low impedeace which are represented by $0$ and $1$. When the forward voltage is larger than the threshold voltage, the impedance will be lowered, and when reverse voltage is larger than the threshold voltage, the impedance will change to high impedance. Fig. ~\ref{fig5} is the implementation of XOR operator using memristor, Table 1 shows the stimulating voltage sequence of the XOR function, where the final output is stored in the memristor $T$. Fig. ~\ref{fig6} gives the general architecture of the memristive edge detector. \cite{Zhou13} has proposed a feasible scheme to produce the edge map by scanning the image horizontally and vertically, and applying the above simulating voltages.
The advantage of embracing this new architecture is that one can build a much more compact circuit with virtually no loss in functionality. However, we believe that an XOR function is probably not the most ideal method for edge detection. We intend to reorganize the circuit structure using memristive latches proposed by \cite{Robinett10}.

\section*{Acknowledgment}
We wish to thank Mr. Yunji Chen for his encouragement and guidance that inspired this project. We would also like to thank Ms. Shengyuan Zhou and Ms. Tian Zhi for her help. None, however, is responsible for any remaining errors.

\bibliographystyle{IEEEtran}

\begin{thebibliography}{1}

\bibitem{Gonzalez}
R.~Gonzalez and R.~Woods, \textit{Digital image processing (3rd Edition)}. Upper Saddle River, NJ, USA: Prentice-Hall, Inc., 2006.

\bibitem{Sobel68}
I.~Sobel and G.~Feldman, ``A 3x3 isotropic gradient operator for image processing," presented at the Stanford Artificial Intelligence Project (SAIL), 1968.

\bibitem{Canny86}
J.~Canny, ``A computational approach to edge detection," \textit{IEEE Trans. Pattern Anal. Mach. Intell.}, vol. 6, pp. 679-698, 1986.

\bibitem{Guo10}
Z.~Guo, W.~Xu, and Z.~Chai, ``Image edge detection based on FPGA," in \textit{Distributed Computing and Applications to Business Engineering and Science (DCABES), 2010 Ninth International Symposium on Distributed Computing and Applications to Business, Engineering and Science}, pp. 169-171. IEEE, 2010.

\bibitem{Abbasi07}
T.~Abbasi and M.~Abbasi, ``A novel FPGA-based architecture for Sobel edge detection operator," \textit{International Journal of Electronics}, vol. 94, no. 9, pp. 889-896, 2007.

\bibitem{Suryakant12}
N.~Suryakant, ``Edge Detection using Fuzzy Logic in MATLAB," International Journal of Advanced Research in Computer Science and Software Engineering, vol. 2, no. 4 (2012).

\bibitem{Zhang15}
L.~Zhang, Y.~Sun, and F.~Chen, ``An improved edge detection algorithm based on fuzzy theory," \textit{2015 12th International Conference on Fuzzy Systems and Knowledge Discovery (FSKD)}. IEEE, 2015.

\bibitem{Bailey}
D.~Bailey, \textit{Design for embedded image processing on FPGAs}. John Wiley \& Sons, 2011.

\bibitem{Merrikh14}
F.~Merrikh-Bayat and S.~Shouraki, ``Memristive fuzzy edge detector", \textit{Journal of Real-Time Image Processing}, vol. 9, no. 3, pp. 479-489, 2014.

\bibitem{Lehtonen09}
E.~Lehtonen and M.~Laiho, ``Stateful implication logic with memristors," in \textit{Proceedings of the 2009 IEEE/ACM International Symposium on Nanoscale Architectures}, pp. 33-36. IEEE Computer Society, 2009.

\bibitem{Zhou13}
J. Zhou, J. Wu, and Y. Tang, ``Edge Detection of Binary Image Based on Memristors," \textit{Advanced Materials Research}, vol. 791. Trans Tech Publications, 2013.

\bibitem{Robinett10}
W.~Robinett \textit{et al.}, ``A memristor-based nonvolatile latch circuit," \textit{Nanotechnology}, vol. 21, no. 23, 235203, 2010.

\bibitem{Pan13}
H.~Pan, ``An Image Edge Detection Algorithm Based on Fuzzy Theory," \textit{J Chongqing Technol Business Univ. (Nat Sci Ed)}, vol. 30, no. 7, pp. 53-56, 2013.

\bibitem{Liu81}
H.~Liu, ``Real-time Sobel Operator Image Edge Detector," \textit{Huazhong University Sc. \& Tech.}, vol.12, no. 1, pp. 15-22, 1981.

\bibitem{Pandey14}
J.~Pandey \textit{et al.}, ``A Novel Architecture for FPGA Implementation of Otsu’s Global Automatic Image Thresholding Algorithm, " \textit{2014 27th International Conference on VLSI Design and 2014 13th International Conference on Embedded Systems}, 2014.
\end{thebibliography}

\end{document}